\documentclass[11pt,a4paper]{article}
\usepackage[hyperref]{emnlp-ijcnlp-2019}
\usepackage{times}
\usepackage{latexsym}

\usepackage{url}

\usepackage{proof}
\usepackage{xspace}
\usepackage{multirow}
\usepackage{forest}
\usepackage{caption}
\usepackage{subcaption}
\usepackage{stix}
\usepackage{rotating}
\usepackage{dsfont}
\usepackage{amsmath}
\usepackage{amssymb}
\usepackage{amsthm}
\usepackage{algorithm}
\usepackage{float}
\usepackage[noend]{algpseudocode}
\definecolor{dollarbill}{rgb}{0.52, 0.73, 0.4}
\definecolor{deepmagenta}{rgb}{0.8, 0.0, 0.8}
\definecolor{coralred}{rgb}{1.0, 0.25, 0.25}

\newcommand{\sentpair}{(\vecs, \vect)\xspace}

\newcommand{\notes}[1]{}%{\it {\small {#1}}}}

\theoremstyle{definition}

\theoremstyle{plain}

\newcommand{\veca}{\ensuremath{\mathbf{a}}\xspace}

\newcommand{\vecs}{\ensuremath{\mathbf{s}}\xspace}
\newcommand{\vect}{\ensuremath{\mathbf{t}}\xspace}
\newcommand{\veco}{\ensuremath{\mathbf{o}}\xspace}

\newcommand{\ith}[1]{\ensuremath{i^{{th}}}}

\newcount\permx
\newcount\permy
\def\permdot#1#2{
\permx=#1 \advance\permx by-1
\permy=#2 \advance\permy by-1
\psframe[fillcolor=black, fillstyle=solid]
(\permx,\permy)(#1, #2)
}

\newcommand{\boxnum}[1]{{\setlength{\fboxsep}{1pt}\raisebox{1pt}{\hspace{1pt}\fbox{\tiny #1}\hspace{1pt}}}}
\newcommand{\ind}[1]{\ensuremath{_{\kern-0.5pt\boxnum{#1}}}}

\newcommand{\smallnt}[1]{\ensuremath{_{\mbox{\tiny PP}}}\xspace}

\newcommand{\pseudocode}{Algorithm}
\floatname{algorithm}{\pseudocode}

\iffalse

\else

\fi

\newcommand{\eos}{\mbox{\scriptsize \texttt{<eos>}}\xspace}

\aclfinalcopy % Uncomment this line for the final submission

\title{
Simpler and Faster Learning of Adaptive Policies \\ for Simultaneous Translation 
}

\author{Baigong Zheng $^{1,}$\thanks{\; These authors contributed equally.} \,
  Renjie Zheng $^{2, \ast}$ \,
  Mingbo Ma $^{1}$ \,
  Liang Huang $^{1,2}$
\\
  $^{1}$Baidu Research, Sunnyvale, CA, USA \\
  $^{2}$Oregon State University, Corvallis, OR, USA \\
  {\small \texttt{\{baigongzheng, mingboma\}@baidu.com \qquad zrenj11@gmail.com} } 
}

\date{}

\begin{document}
\maketitle
\begin{abstract}
  %Simultaneous translation is widely useful but remains challenging.  Previous work falls into two main categories: (a) fixed-latency policies (such as the wait-$k$ policy in \namecite{ma+:2018}), which, due to the difference between source and target word order, have to aggressively hallucinate future content; and (b) adaptive policies learned by reinforcement learning, which do not anticipate, but suffer from unstable training due to randomness in exploration.  To combine the merits of both approaches, we propose a simple supervised-learning framework to learn an adaptive policy from oracle READ/WRITE sequences generated from parallel text.  At each step, such an oracle sequence chooses to WRITE the next target word if the available source sentence context provides enough information to do so, otherwise READ the next source word.  German$\leftrightarrow$English experiments show that our method can learn flexible policies with better BLEU scores and similar latencies compared to previous work, without retraining the underlying NMT model.
  Simultaneous translation is widely useful but remains challenging.  Previous work falls into two main categories: (a) fixed-latency policies such as~\citet{ma+:2019}
  and (b) adaptive policies such as~\citet{gu+:2017}.
  The former are simple and effective, but %inevitably
  have to aggressively predict future content
  due to diverging source-target word order; 
  the latter do not anticipate, but suffer from unstable and inefficient training.  To combine the merits of both approaches, we propose a simple supervised-learning framework to learn an adaptive policy from oracle READ/WRITE sequences generated from parallel text.  At each step, such an oracle sequence chooses to WRITE the next target word if the available source sentence context provides enough information to do so, otherwise READ the next source word.  Experiments on German$\leftrightarrow$English show that our method, without retraining the underlying NMT model, can learn flexible policies with better BLEU scores and similar latencies compared to previous work.

\end{abstract}

% !TEX root = main.tex

\section{Introduction}
Simultaneous translation outputs target words while the source sentence is being received, and is widely useful in international conferences, negotiations and press releases.
However, although there is significant progress in machine translation (MT) recently, simultaneous machine translation is still one of the most challenging tasks. This is because it is hard to balance translation quality and latency, especially for the syntactically divergent language pairs, such as English and Japanese.

Researchers previously study simultaneous translation as a part of real-time speech-to-speech translation system~\cite{mahsa+:2013, bangalore+:2012, fugen+:2007, sridhar+:2013, jaitly2016online, graves2013speech}.  
Recent simultaneous translation research focuses on obtaining a strategy, called a {\em policy}, to decide whether to wait for another source word (READ action) or emit a target word (WRITE action).
The obtained policies fall into two main categories: (1) fixed-latency policies~\cite{ma+:2019, dalvi+:2018} and (2) context-dependent adaptive policies~\cite{grissom+:2014, cho+:16, gu+:2017, alinejad+:2018, arivazhagan+:2019, zheng+:2019a}.  
As an example of fixed-latency policies, 
wait-$k$~\cite{ma+:2019} starts by waiting for the first few source words and then outputs one target word after receiving each new source word until the source sentence ends. It is easy to see that this kind of policy will inevitably need to guess the future content, which can often be incorrect. 
Thus, an adaptive policy (see Table~\ref{tab:ex} as an example), which can decides on the fly whether to take READ action or WRITE action, is more desirable for simultaneous translation. 
Moreover, the widely-used beam search technique become non-trivial for fixed policies~\cite{zheng+:2019b}.
%Previous work~\cite{gu+:2017, alinejad+:2018} depends on reinforcement learning (RL) method to learn such an adaptive policy. Their training process is very unstable and inefficient due to its exploration process.
%%Furthermore, such a learned policy cannot control latency in applications, which reduces its applicability in many scenarios.
%Furthermore, such {\em one} learned policy cannot adapt its latency after training process, and we will need to train multiple policies for scenarios with different latency requirements.

\begin{table*}[h]
\centering
\resizebox{1\textwidth}{!}{%
\setlength{\tabcolsep}{3pt}
\begin{tabular}{c | c c c c c c c c c c c c c c c c c c c c c c c c}
\hline
German &             & Ich &   &  & bin & & mit & & dem & & Bus &    &   &      & & nach & & Ulm & &  gekommen & & &  & \\
\hline
Gloss &              & I   &   &  & am  & & with& & the & & bus &    &   &      & & to   & & Ulm & &  come & & & & \\
\hline
Action & R    &     & W &R &     &R&     &R&     &R&     &W   & W & W    &R&      &R&     &R&       & W & W & W & W \\
\hline
Translation &        &     & I &  &     & &     & &     & &     &took& the & bus& &      & &     & &       &to & come & to & Ulm \\
\hline
\end{tabular}
}
\vspace{-5pt}
\caption{An example for READ/WRITE action sequence. R represents READ and W represents WRITE.}
\label{tab:ex}
\vspace{-15pt}
\end{table*}

To represent an adaptive policy, previous work shows three different ways: (1) a rule-based decoding algorithm~\cite{cho+:16}, (2) an original MT model with an extended vocabulary~\cite{zheng+:2019a}, and (3) a separate policy model~\cite{grissom+:2014, gu+:2017, alinejad+:2018, arivazhagan+:2019}.  
The decoding algorithm~\cite{cho+:16} applies heuristics measures and does not exploit information in the hidden representation, while the MT model with an extended vocabulary~\cite{zheng+:2019a} needs guidance from a restricted dynamic oracle to learn an adaptive policy, whose size is exponentially large so that approximation is needed.
A separate policy model could avoid these issues.
However, previous policy-learning methods either depends on reinforcement learning (RL)~\cite{grissom+:2014, gu+:2017, alinejad+:2018}, which makes the training process unstable and inefficient due to exploration, or applies advanced attention mechanisms~\cite{arivazhagan+:2019}, which requires its training process to be autoregressive, and hence inefficient.
Furthermore, each such learned policy cannot change its behaviour according to different latency requirements at testing time, and we will need to train multiple policy models for scenarios with different latency requirements.

To combine the merits of fixed and adaptive policies, and to resolve the mentioned drawbacks, we propose a simple supervised learning framework to learn an adaptive policy, and show how to apply it with controllable latency. This framework is based on sequences of READ/WRITE actions for parallel sentence pairs, so we present a simple method to generate such  an action sequence for each sentence pair with a pre-trained neural machine translation (NMT) model.\footnote{Previous work~\cite{he+:2015, niehues+:2018} shows that carefully generated parallel training data can help improve simultaneous MT or low-latency speech translation. This is different from our work in that we generates action sequences for training policy model instead of parallel translation data for MT model.}
Our experiments on German$\leftrightarrow$English dataset show that our method, without retraining the underlying NMT model, leads to better policies than previous methods, and achieves better BLEU scores than (the retrained) wait-$k$ models at low latency scenarios.

% !TEX root = main.tex

\section{Generating Action Sequences}
In this section, we show how to generate action sequences for parallel text.
Our simultaneous translation policy can take two actions: READ (receive a new source word) and WRITE (output a new target word). A sequence of such actions for a sentence pair $\sentpair$ defines one way to translate $\vecs$ into $\vect$. Thus, such a sequence must have $|\vect|$ number of WRITE actions.
However, not every action sequence is good for simultaneous translation. For instance, a sequence without any READ action will not provide any source information, while a sequence with all $|\vecs|$ number of READ actions before all WRITE actions usually has large latency. Thus the ideal sequences for simultaneous translation should have the following two properties: 

\begin{itemize}
\setlength\itemsep{-2pt}
\item there is no anticipation during translation, i.e. when choosing WRITE action, there is enough source information for the MT model to generate the correct target word;
\item the latency is as low as possible, i.e. the WRITE action for each target word appears as early as possible.
\end{itemize} 
Table~\ref{tab:ex} gives an example for such a sequence.

\begin{algorithm}
\caption{Generating Action Sequence}
\label{alg:rw}
\begin{algorithmic}
\State {\bf Input}: sentence pair $\sentpair$, integer $r$, model $M$
\State $id_s \leftarrow 1$,  $id_t \leftarrow 1$
\State Seq $\leftarrow [R]$
\While {$id_t \le |\vect|$}
\If {$rank_M(t_{id_t} | \vecs_{\le id_s}) \le r$ or $id_s = |\vecs|$}
\State Seq $\leftarrow$ Seq $+ [W]$
\State $id_t \leftarrow id_t + 1$
\Else 
\State Seq $\leftarrow$ Seq $+ [R]$
\State $id_s \leftarrow id_s + 1$
\EndIf
\EndWhile
\Return Seq
\end{algorithmic}
\end{algorithm}

In the following, we present a simple method to generate such an action sequence for a sentence pair $\sentpair$ using a pre-trained NMT model, assuming this model can make reasonable prediction given incomplete source sentence.
Our method is based on this observation: {\em if the rank of the next ground-truth target word is high enough in the prediction of the model, then this implies that there is enough source-side information for the model to make a correct prediction.}
Specifically, we sequentially input the source words to the pre-trained model, and use it to predict next target word. If the rank of the gold target word is high enough, we will append a WRITE action to the sequence and then try the next target word; otherwise, we append a READ action and input a new source word. 
Let $r$ be a positive integer, $M$ be a pre-trained NMT model, $\vecs_{\le i}$ be the source sequence consisting of the first $i$ words of $\vecs$,  and $rank_M(t_j | \vecs_{\le i})$ be the rank of the target word $t_j$ in the prediction of model $M$ given sequence $\vecs_{\le i}$.  Then the generating process can be summarized as Algorithm~\ref{alg:rw}.  

Although we can generate action sequences balancing the two wanted properties with appropriate value of parameter $r$, the latency of generated action sequence may still be large due to the word order difference between the two sentences.  
To avoid this issue, we filter the generated sequences with the latency metric {\em Average Lagging} (AL) proposed by~\citet{ma+:2019}, which quantifies the latency in terms of the number of source words and avoids some flaws of other metrics like {\em Average Proportion} (AP)~\cite{cho+:16} and {\em Consecutive Wait} (CW)~\cite{gu+:2017}.
Another issue we observed is that, the pre-trained model may be too aggressive for some sentence pair, meaning that it may write all target words without seeing the whole source sentence. This may be because the model is also trained on the same dataset. To overcome this, we only keep the action sequences that will receive all the source words before the last WRITE action.
After the filtering process, each action sequence has AL less than a fixed constant $\alpha$ and receives all source words. 
%And these sequences will be used in our supervised learning algorithm.

% !TEX root = main.tex
\section{Supervised-Learning Framework for Simultaneous Translation Policy}
Given a sentence pair and an action sequence for this pair, we can apply supervised learning method to learn a parameterized policy for simultaneous translation. For the policy to be able to choose the correct action, its input should include information from both source and target sides. Since we use Transformer~\cite{vaswani+:2017} as our underlying NMT model in this work, we need to recompute encoder hidden states for all previous seen source words, which is the same as done for wait-$k$ model training~\cite{ma+:2019}.\footnote{In our experiments, the decoder and policy model combined need about 0.0445 seconds on average to generate one target word (this might include multiple READ's), while it takes on average only about 0.0058 seconds to recompute all encoder states for each new source word. So we think this re-computation might not be a serious issue for system efficiency.}  
The policy input $o_i$ at step $i$ consists of three components from this model: 
\vspace{-5pt}
\begin{itemize}
\setlength\itemsep{-2pt}
\item $h^s_i$: the last-layer hidden state from the encoder for the first source word at step $i$;
\item $h^t_i$: the last-layer hidden state from the decoder for the first target word at step $i$;\footnote{The hidden states of the first target word will be re-computed at each step.}
\item $c_i$: cross-attention scores at step $i$ for the current input target word on all attention layers in decoder, averaged over all current source words.
\end{itemize} 
\vspace{-6pt}
That is $ o_i = [h^s_i, h^t_i, c_i]$. 
%Since Transformer encoder (decoder) has multiple self-attention layers, the last hidden state for the first source (target) word has included information from all current source (target) words\footnote{We also tried the hidden states for the current last source and target words, but did not get any improvement.}. And the averaged cross-attention  scores indicate the correlation between current source sentence and predicted target word. 

Let $a_i$ be the $i$-th action in the given action sequence $\veca$. Then the decision of our policy on the $i$-th step depends on all previous inputs $\veco_{\le i}$ and all taken actions $\veca_{< i}$.
We want to maximize the probability of the next action $a_i$ given those information:
\vspace{-4pt}
\[
\vspace{-4pt}
\max p_{\theta} (a_i | \veco_{\le i}, \veca_{< i})
\]
where $p_{\theta}$ is the action distribution of our policy parameterized by $\theta$.

% !TEX root = main.tex
\section{Decoding with Controllable Latency}

To apply the learned policy for simultaneous translation, we can choose at each step the action with higher probability . However, different scenarios may have different latency requirements. Thus, this greedy policy may not always be the best choice for all situations. Here we present a simple way to implicitly control the latency of the learned policy without retraining the policy model.

Let $\rho$ be a probability threshold. For each step in translation, we choose READ action only if the probability of READ is greater than $\rho$; otherwise we choose WRITE action. Thus, this threshold balances the tradeoff between latency and translation quality: with larger $\rho$, the policy prefers to take WRITE actions, providing lower latency; and with smaller $\rho$, the policy prefers to take READ actions, and provides more conservative translation with larger latency.

% !TEX root = main.tex

\begin{figure}[h]
\centering
\vspace{-5pt}
\includegraphics[width=.9\linewidth]{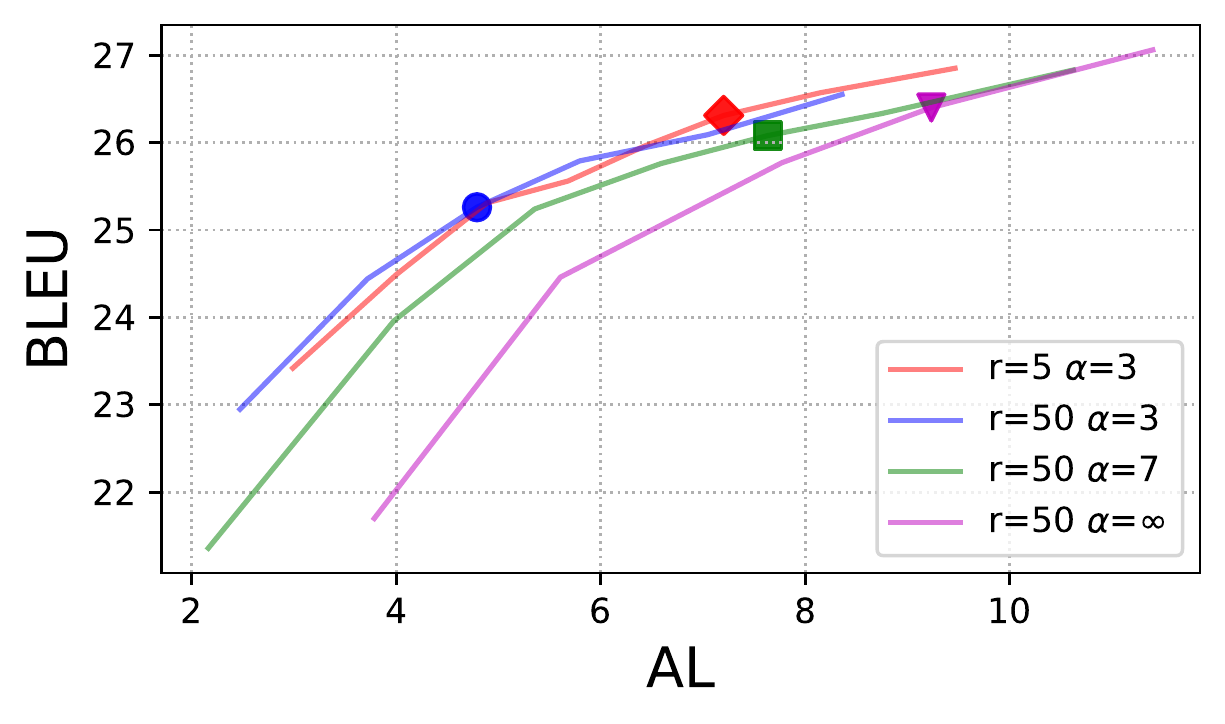}
\vspace{-10pt}
\caption{
Translation quality against latency on DE$\to$EN dev set. 
Lines are obtained with different probability thresholds $\rho$.
Markers represent %the models with
$\rho = 0.5$.
%\textcolor{coralred}{$\mdlgblkdiamond$}: wait-$k$ models for $k \in \{1, 2, 3, 4, 5, 6, 7\}$,
%\textcolor{dollarbill}{$\mdlgblkcircle$}: test-time wait-$k$ for $k \in \{1, 2, 3, 4, 5, 6, 7\}$,
%\textcolor{dollarbill}{$\bigstar$}: full-sentence translation with pre-trained NMT model.
%Blue shapes represent our SL models trained with action sequences generated using different parameters (
%\textcolor{blue}{$\smallblacktriangleright$}: $r=50$ and $\alpha=3$, 
%\textcolor{blue}{$\blacksquare$}: $r=50$ and $\alpha=5$, 
%\textcolor{blue}{$\blacktriangle$}: $r=50$ and $\alpha=7$, 
%\textcolor{blue}{$\blacktriangledown$}: $r=50$ and $\alpha=\infty$, 
%\textcolor{blue}{$\smallblacktriangleleft$}: $r=100$ and $\alpha=3$, 
%\textcolor{blue}{$\blacklozenge$}: $r=5$ and $\alpha=3$).
}
\label{fig:param}
\vspace{-10pt}
\end{figure}

\begin{figure*}[!t]
%\centering
%\hspace{-30pt}
%\begin{minipage}[h]{.45\textwidth}
%\begin{subfigure}[h]{.9\textwidth}
%\centering
\includegraphics[width=0.49\linewidth]{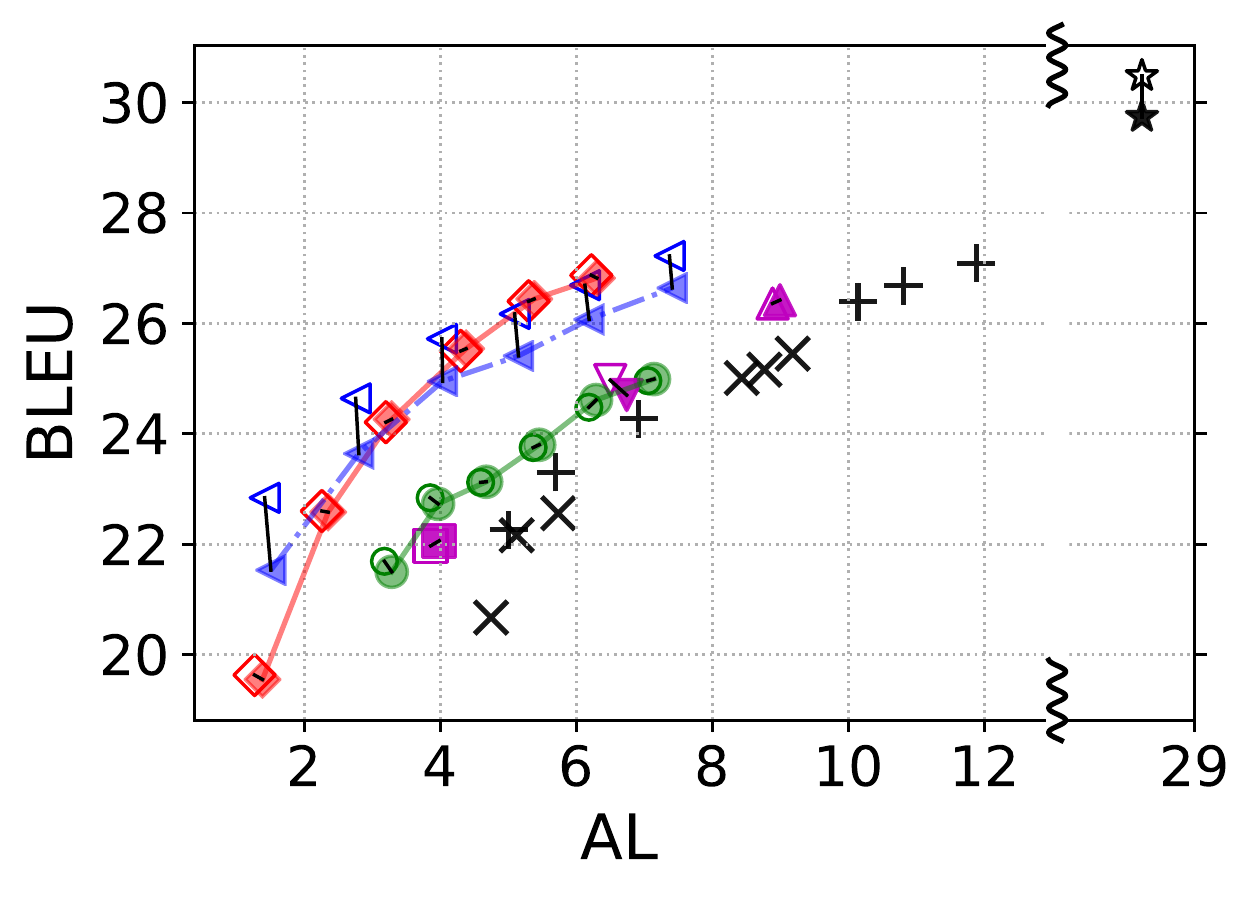}
%\vspace{-20pt}
%\hspace{30pt}
%\caption{\label{fig:a} DE$\to$EN}
%\end{subfigure}
%\end{minipage}\qquad
%\begin{minipage}[h]{.45\textwidth}
%\begin{subfigure}[h]{.9\textwidth}
%\centering
%\hspace{-30pt}
\ \ \ 
\includegraphics[width=0.49\linewidth]{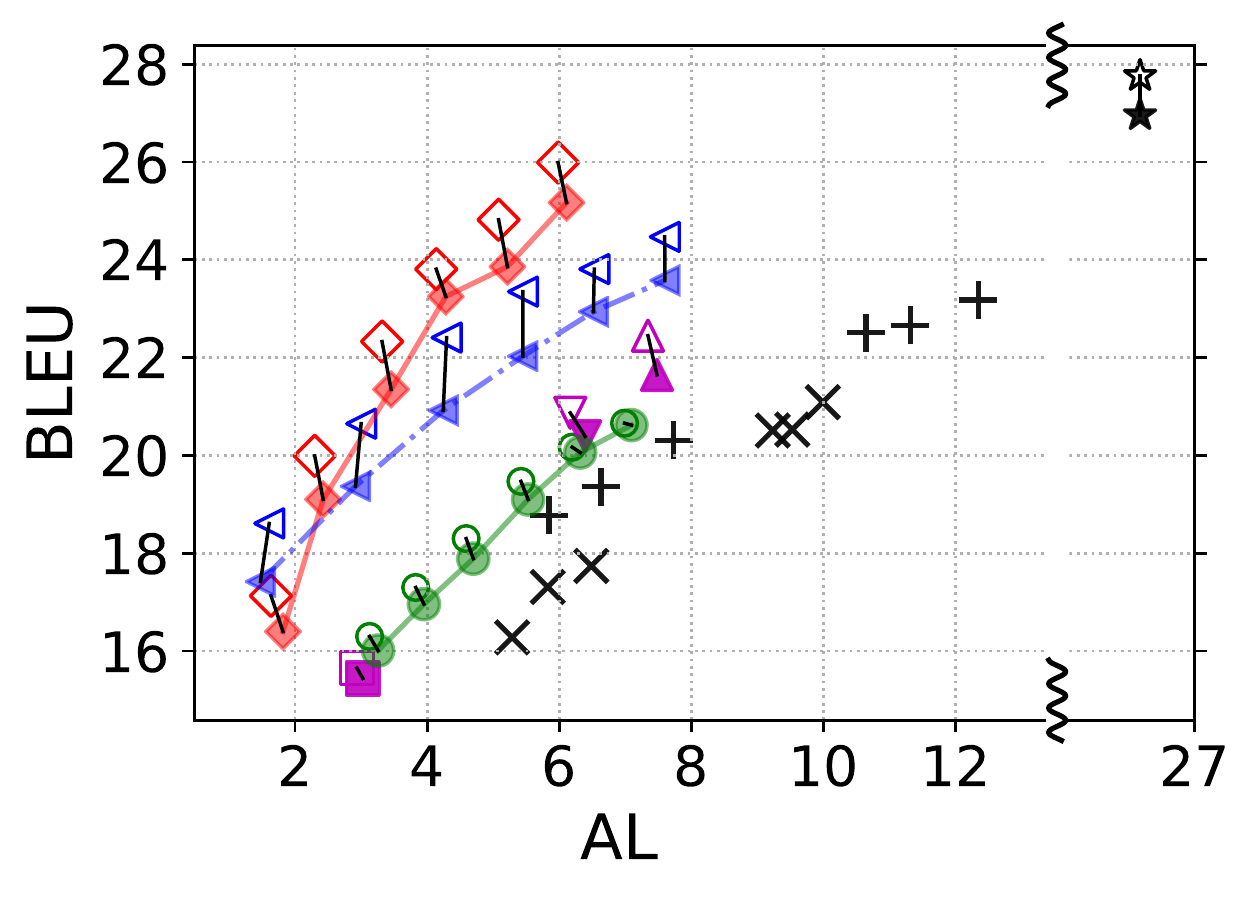}
%\vspace{-20pt}
%\caption{\label{fig:b} EN$\to$DE}
%\end{subfigure}
%\end{minipage}
\vspace{-20pt}
\caption{
Comparing performances of different methods on testing sets.
Left: DE$\to$EN. Right: EN$\to$DE.
The shown pairs are results of greedy decoding (solid shapes) and beam search (empty shapes, beam-size = 5).  
\textcolor{coralred}{$\mdlgblkdiamond$}: wait-$k$ models for $k \in \{1, 2, 3, 4, 5, 6 \}$,
\textcolor{dollarbill}{$\mdlgblkcircle$}: test-time wait-$k$ for $k \in \{1, 2, 3, 4, 5, 6\}$,
\textcolor{blue}{$\smallblacktriangleleft$}: our SL model with threshold $\rho \in \{ 0.65, 0.6, 0.55, 0.5, 0.45, 0.4\}$,
\textcolor{black}{$\bigstar$}: full-sentence translation model,
\textcolor{deepmagenta}{$\blacksquare$}: RL with CW = 2,
\textcolor{deepmagenta}{$\blacktriangledown$}: RL with CW = 5,
\textcolor{deepmagenta}{$\blacktriangle$}: RL with CW = 8,
$\times$: WID for $s_0 \in \{ 2, 4, 6\}$ and $\delta \in \{2, 4\}$,
$+$: WIW for $s_0 \in \{2, 4, 6\}$ and $\delta \in \{1, 2\}$.
}
\label{fig:pfm}
\vspace{-8pt}
\end{figure*}

\section{Experiments}

\paragraph{Dataset}

We conduct experiments on English$\leftrightarrow$German (EN$\leftrightarrow$DE) simultaneous translation. 
We use the parallel corpora  from WMT 15  for training,
newstest-2013 for validation
and newstest-2015 for testing.\footnote{\href{http://www.statmt.org/wmt15/translation-task.html}{\scriptsize\tt http://www.statmt.org/wmt15/translation-task.html}}
All datasets are tokenized and segmented into sub-word units with byte-pair encoding (BPE)~\cite{sennrich+:2016},
and we only use the sentence pairs of lengths less than 50 (on both sides) for training.

\paragraph{Model Configuration}
We use Transformer-base \cite{vaswani+:2017} as our NMT model %, which have the same architecture as the base model in the original paper,
and our implementation is based on PyTorch-based OpenNMT~\cite{Klein+:2017}.
We add an  \eos token on the source side, which is not included in the original OpenNMT codebase.
Our recurrent policy model consists of one GRU layer with 512 units, one fully-connected layer of dimension 64 followed by ReLU activation, and one fully-connected layer of dimension 2 followed by a softmax function to produce the action distribution.  
%For policy training, we use optimizer Adam~\cite{kingma+:2014} with the initial learning rate of 0.0001, which is halved every 200 steps after the first 1000 steps.
%We train our policy model with a batch size of 32, 
%In the following,
We use BLEU \cite{BLEU:2002} as the translation quality metric
and  Averaged Lagging (AL)~\cite{ma+:2019} as the latency metric.%, which avoids some limitations from other metrics.

\paragraph{Effects of Generated Action Sequences}
We first analyze the effects of the two parameters in the generation process of action sequences: the rank $r$ and the filtering latency $\alpha$. We fix $\alpha = 3$ and choose the rank $r \in \{5, 50\}$; then we fix $r=50$ and choose the latency $\alpha \in \{3, 7, \infty\}$ to generate action sequences on DE$\to$EN direction.%, whose statistical information is provided in the supplemental material.
 Figure~\ref{fig:param} shows the performances of resulting models with different probability thresholds $\rho$.
We find that smaller $\alpha$ helps achieve better performance and our model is not very sensitive to values of rank.  Therefore, in the following experiments, we report results with $r = 50$ and $\alpha = 3$.

% example
\begin{table*}[!t]
\resizebox{0.99\linewidth}{!}{%
\centering
\setlength{\tabcolsep}{-3pt}
\begin{tabular}{ c | c c c c c c c c c c c c c c c c c c c c c c c c c c c c c c c c c c}
\hline
German  & \ \ die & & \ \ deutsche  & & bahn & & will & &im & & \kern-0.9em kommenden \kern-0.8em& &jahr & &\kern-0.3em die & &kin- & &\kern-0.9em zi- \kern-0.9em& &g- \ \  & & tal-& & -- & & \kern-0.5em bahn- & & \kern-0.5em strecke \kern-0.5em  & &\kern-0.5em verbessern \kern-0.5em  & & . \ \ \ \ \ & \\ [5pt]
\hline
gloss & \ \ the  & &\ \  German \! \! \! & &train &         & want & &\kern-0.4em in the \kern-0.5em & &\kern-0.9em coming\kern-0.8em & &year & & \kern-0.3em the  & & & &\kern-0.9em Kinzigtal\kern-0.9em & & & & & & -- & & & \kern-0.9em railroad track \kern-0.9em  & & &\kern-0.5em improve & & . \ \ \ \ \ & \\ [5pt]
\hline
wait-3   &     & &        & &      &deutsche & & bahn & & wants & & to & & make & &the & &cinema & &show & &a & &success & &this & &coming & &year & &. & & \\ [5pt]
\hline
wait-5   &     & &        & &      &  & & & &deutsche & &bahn & &wants & &to & &introduce & &the & &kin- & &zi- & &g- & &tal & &railway & &line & & next year . \\ [5pt]
\hline
test-time \ \  &  & & & & &deutsche  & & bahn& &wants & &the & &german & &railways & &to & &be & & the& &kin- & &z- & &ig- & &tal & &railway & & line in the  \\ 
wait-3 \ \  &  & & & & && & & && && && && && && & & && && && && && & coming year . \\ [5pt]
\hline
test-time  &  & & & & &  & & & &the & &german & &railways & &wants & &to & &take & &the & &german & &train & &to & &the & &german & & railways in the \\ 
wait-5    &  & & & & &  & & & && && && && && && && && && && && && & coming year . \\ [5pt]
\hline
SL &  &the & &\kern-0.5em german & &railway  & &wants the & & & & & & & & & & & & & & & & & & & & & & & & & & kinzigtal railway \\
 policy &  & & & & && && & & & & & & & & & & & & & & & & & & & & & & & & & to be improved \\ 
 &  & & & & && && & & & & & & & & & & & & & & & & & & & & & & & & & next year . \\ [5pt]
\hline
RL &  &    & &       & &         & &the german \kern-0.5em & & & & & &railways wants & & & & & &the german & & & & & &railway will \kern-0.5em & & & & & &\kern-0.5em improve the & & kinzigtal railway \\
 policy &  &    & &       & &         & && & & & & && & & & & && & & & & && & & & & & & & next year . \\ [5pt]
\hline
\end{tabular}
}
\caption{German-to-English example from validation set.}
\label{tab:ex2}
\vspace{-10pt}
\end{table*}

\paragraph{Performance Comparison}
We compare our method on EN$\leftrightarrow$DE directions with different methods: greedy decoding algorithms Wait-If-Worse/Wait-If-Diff (WIW/WID) of~\citet{cho+:16}, RL method of~\citet{gu+:2017}, wait-$k$ models and test-time wait-$k$ methods of~\citet{ma+:2019}.  
Both of WIW and WID only use the pre-trained NMT model, and the algorithm initially reads $s_0$ number of source words and chooses READ only if the probability of the most likely target word decreases when given another $\delta$ number of source words (WIW) or the most likely target word changes when given another $\delta$ number of source words (WID).
For RL method, we use the same kind of input and architecture for policy model.\footnote{We only report results obtained with CW-based reward functions for they outperform those with AP-based ones.}
Test-time wait-$k$ means decoding with wait-$k$ policy using the pre-trained NMT model.  All methods share the same underlying pre-trained NMT model except for the wait-$k$ method,
which retrains the NMT model from scratch, but with the same architecture as the pre-trained model.

Figure~\ref{fig:pfm} shows the performance comparison.
Our models on both directions can achieve higher BLEU scores with the similar latency than WIW, WID, RL model and test-time wait-$k$ method, implying that our method learns a better policy model than the other methods when the underlying NMT model is not retrained.\footnote{Our test-time wait-$k$ results are better than those in~\citet{ma+:2019}, because the added source side \eos token helps the full-sentence model to learn when the source sentence ends. Without this token, test-time wait-$k$ generates either very short target sentences or many punctuations for small $k$'s.} Compared with wait-$k$ models, our models with beam search achieve higher BLEU scores when latency AL is small, which we think will be the most useful scenarios of simultaneous translation. 
Furthermore, this figure also shows that our model can achieve good performance on different latency conditions by controlling the threshold $\rho$, so we do not need to train multiple models for different latency requirements.
%We also provide a translation example in the supplemental material to compare different methods.
We also provide a translation example in Table~\ref{tab:ex2} to compare different methods.
%\paragraph{Example}
%In Table~\ref{tab:ex2}, we provide one example from validation set on DE$\to$EN direction to compare the different methods.  This example shows that our policy model can generate correct translation with low latency if possible (shown in the beginning part of the example).  However, when there is not enough information, it can choose to wait for more information to avoid risky anticipation. This is shown by waiting for the last verb in the example.  Models from other methods either make wrong anticipation or generate repeated translation to wait for new information.

\begin{figure}[t]
\centering 
%\vspace{-5pt}
\includegraphics[width=.97\linewidth]{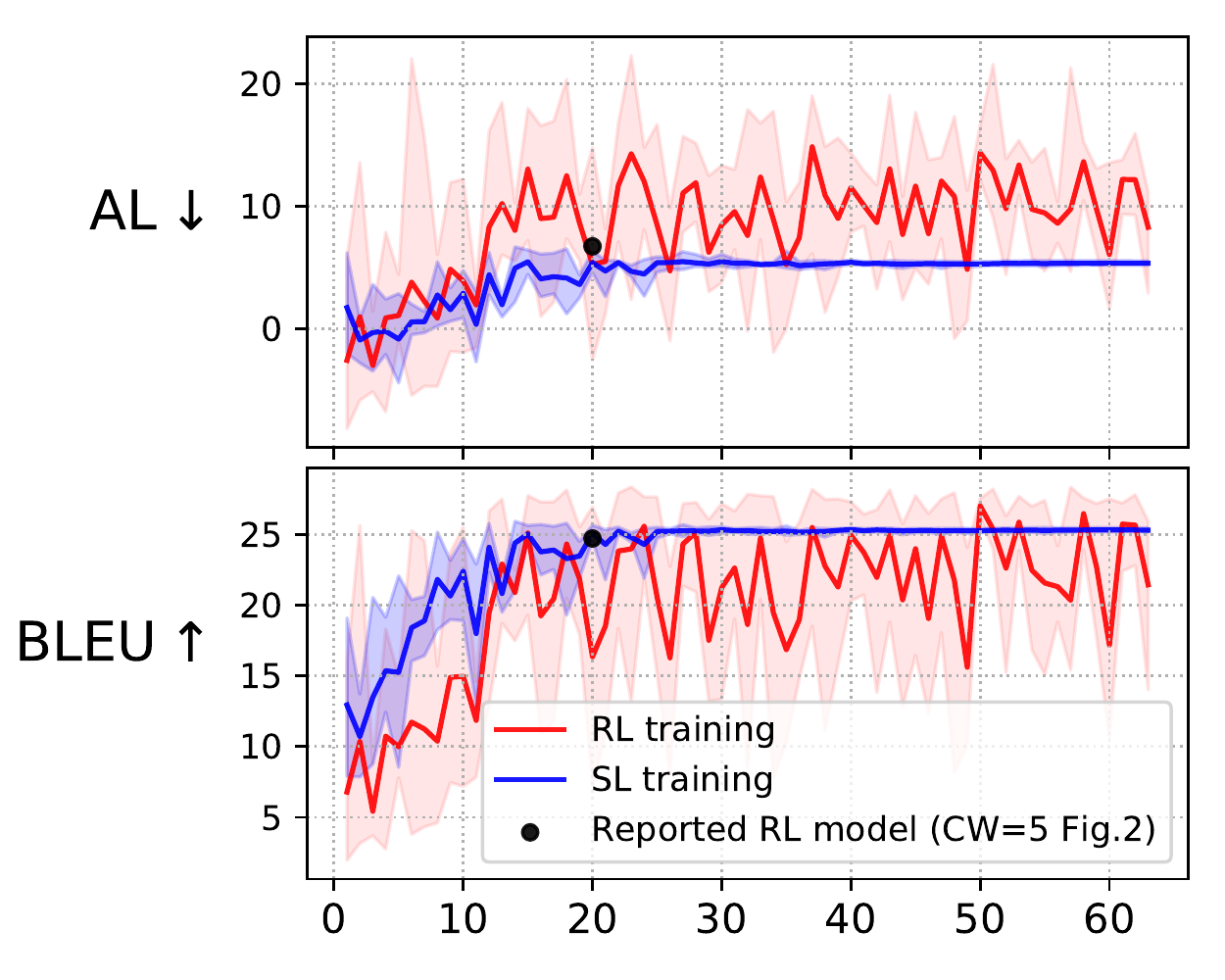}
\vspace{-10pt}
\caption{
Learning curves averaged over four independent training runs on DE$\to$EN direction.
The $x$-axis represents training steps ($\times$ 50).
}\label{fig:lc}
\vspace{-5pt}
\end{figure}

\paragraph{Learning Process Analysis}
We analyze two aspects of the learning processes of different methods: stability and training time.
Figure~\ref{fig:lc} shows the learning curves of the training processes of RL method and our SL method, averaged over four different runs with different random seeds on DE$\to$EN direction.
We can see that the training process of our method is more stable and converges faster than the RL method.
Although there are some steps where the RL training process can achieve better BLEU scores than our SL method, the corresponding latencies are usually very big, which are not appropriate for simultaneous translation. We present the training time of different methods in Table~\ref{tab:time}. %\footnote{Since RL training is unstable, we report the training time needed to obtain the BLEU scores reported.} 
 Our method only need about 12 hours to train the policy model with 1 GPU, while the wait-$k$ method needs more than 600 hours with 8 GPUs to finish the training process, showing that our method is very efficient.
Note that this table does not include the time needed to generate action sequences.
The time for this process could be very flexible since we can parallelize this by dividing the training data into separating parts.  In our experiments, we need about 2 hours to generate all action sequences in parallel.

%run time table
\begin{table}[t]
\centering
\resizebox{.48\textwidth}{!}{%
\begin{tabular}{|c  ||  c|  c ||  c|  c|  c|}
\hline
\multirow{2}{*}{DE$\to$EN}& \multicolumn{2}{|c||}{pre-train: 64}  & wait-1 & wait-3 & wait-5 \\
\cline{2-6}
 & SL: +12 & RL: +14.3 & 938 & 966 & 945 \\
\hline
\multirow{2}{*}{EN$\to$DE}& \multicolumn{2}{|c||}{pre-train: 68}  & wait-1 & wait-3 & wait-5 \\
\cline{2-6}
 & SL: +11 & RL: +11.2 & 665 & 665 & 630 \\ 
\hline
\end{tabular}
}
\caption{Training time (in hours) of different methods.  Wait-$k$ training uses 8 GPUs, while others use 1 GPU. The RL time is the average time needed to obtain the three RL models in Figure~\ref{fig:pfm}.
%The "SL" time includes the time we need to generate action sequences in parallel, which is about 2 hours in our experiments.
}
\label{tab:time}
\vspace{-10pt}
\end{table}

\section{Conclusions}
We have proposed a simple supervised-learning framework to learn an adaptive policy based on generated action sequences for simultaneous translation, which leads to faster training and better policies than previous methods, without the need to retrain the underlying NMT model. 
% To apply this framework, we presented one way to generate low-latency action sequences from parallel text. 
\section*{Acknowledgments}
{
We thank Hairong Liu for helpful discussion, Kaibo Liu for helping training baseline models, and the anonymous reviewers for suggestions. We also thank Kaibo Liu for making the AL script available at \url{https://github.com/SimulTrans-demo/STACL}.
}

%\newpage

\bibliography{main}
\bibliographystyle{acl_natbib}

%\newpage

%\appendix
%\input{appendix}

\end{document}